\pdfoutput=1
\documentclass[11pt]{article}

\usepackage[preprint]{acl_latex/acl}

\usepackage{times}
\usepackage{latexsym}

\usepackage[T1]{fontenc}
\usepackage[utf8]{inputenc}

\usepackage{microtype}

\usepackage{inconsolata}

\usepackage{graphicx}

\usepackage{amsmath}
\usepackage{amssymb}
\usepackage{mathtools}
\usepackage{amsthm}

\usepackage{microtype}
\usepackage{graphicx}
\usepackage{subfigure}
\usepackage{booktabs} % for professional tables
\usepackage{soul,enumitem}
\usepackage{algorithm}
\usepackage{algpseudocode}
\usepackage{multirow}
\title{Segment-Based Attention Masking for GPTs}

\author{Shahar Katz$^{*1}$ ~~~~~ Liran Ringel$^{*2}$ ~~~~~ Yaniv Romano$^{2,3}$ ~~~~~ Lior Wolf$^{1}$\\
$^1$Blavatnik School of Computer Science, Tel Aviv University\\
$^2$Department of Computer Science, Technion -- Israel Institute of Technology\\
$^3$Department of Electrical and Computer Engineering, Technion -- Israel Institute of Technology\\
\small{\texttt{\{shaharkatz3@mail,wolf@cs\}.tau.ac.il}}, \small{\texttt{\{liranringel@cs.,yromano@\}technion.ac.il}}
}

\begin{document}
\maketitle
\def\thefootnote{*}\footnotetext{Equal contribution.}
\begin{abstract}
Modern Language Models (LMs) owe much of their success to masked causal attention, the backbone of Generative Pre-Trained Transformer (GPT) models.
Although GPTs can process the entire user prompt at once, the causal masking is applied to all input tokens step-by-step, mimicking the generation process. This imposes an unnecessary constraint during the initial ``prefill'' phase when the model processes the input prompt and generates the internal representations before producing any output tokens.
In this work, attention is masked based on the known block structure at the prefill phase, followed by the conventional token-by-token autoregressive process after that. For example, in a typical chat prompt, the system prompt is treated as one block, and the user prompt as the next one. Each of these is treated as a unit for the purpose of masking, such that the first tokens in each block can access the subsequent tokens in a non-causal manner. Then, the model answer is generated in the conventional causal manner. 
The Segment-by-Segment scheme, illustrated in \autoref{fig:teaser}, entails no additional computational overhead. 
When integrating it into models such as Llama and Qwen, state-of-the-art performance is consistently achieved.
Our code will be available at: \url{https://github.com/shacharKZ/MAS-Segment-Based-Attention-Masking}  .
\end{abstract}

\begin{figure}[h!]
    \centering{\includegraphics[width=\columnwidth]{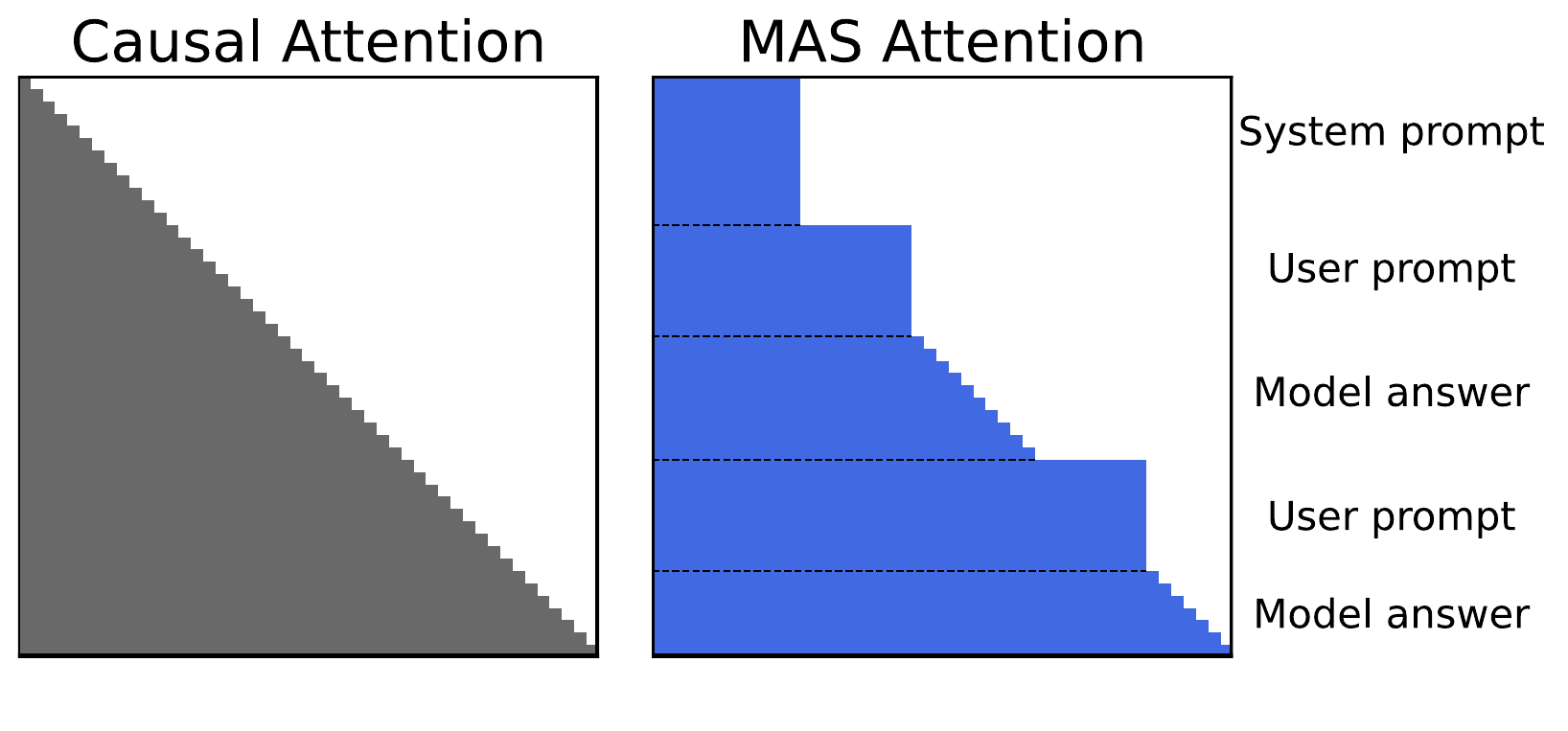}}
    \caption{Causal and MAS attention. The plot shows binary values, where the y-axis represents the index of the current token, and the x-axis represents the set of indices of tokens it can attend to. MAS is inspired by the observation that input prompts are provided to the model as a whole, so they can be masked together in blocks, allowing access to future tokens within the same block of the prompt.\label{fig:teaser}}
    \smallskip
    \smallskip
    \centering{\includegraphics[width=\columnwidth]{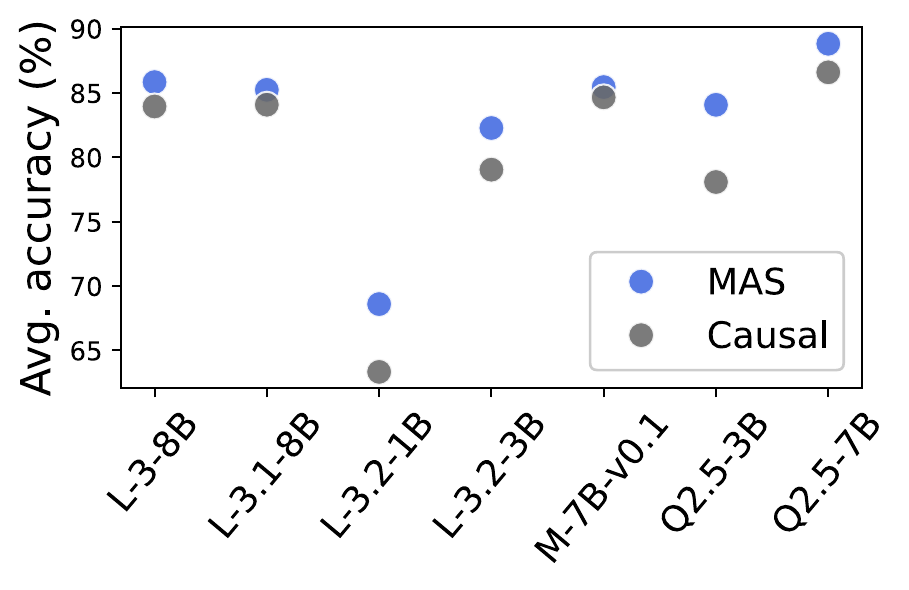}}
    \vspace{-4pt}
  \caption{Model performance on the Commonsense Reasoning benchmark for seven LLMs.
  {L,M,Q stand for Llama, Mistral and Qwen respectively.}}
  \label{fig:summaryresults} 
  \label{MAS overview}
\end{figure}

\section{Introduction}
\label{Introduction}

The introduction of the transformer architecture \cite{vaswani2017attention} has significantly advanced the field of natural language processing (NLP).
Encoder transformer models \cite{devlin2018bert} read text bidirectionally, leveraging both preceding and subsequent words to build a rich contextual representation of the input.
In contrast, decoder models, commonly referred to as GPT models \cite{radfordimproving}, process text unidirectionally, from left to right. This unidirectional structure enables scalability and makes GPTs particularly effective for autoregressive tasks, such as conversational AI.  

The original Transformer architecture introduced by \citet{vaswani2017attention} utilized an encoder-decoder framework, where the encoder built a context for the input, and the decoder generated the output. However, this design requires approximately twice the number of parameters compared to decoder-only models with equivalent capacity.
Efforts such as those by \citet{dong2019unified}, \citet{raffel2020exploring}, and \citet{tayul2} explored unified architectures, where a model's parameters are trained from scratch to function as both encoder and decoder. Despite their potential efficiency, these approaches failed to gain widespread adoption.
In contrast, the remarkable success of decoder-only models, exemplified by GPT-3 \cite{brown2020language}, has shifted the field's focus toward architectures almost exclusively based on causal attention.

While the popularity of GPT models continues to grow, a key limitation is their inability to fully leverage information from future tokens.
Moreover, the community's reliance on established GPT frameworks presents a significant challenge to adopting new attention mechanisms.
For instance, solutions often build on existing models to gain traction, like chain-of-thought prompting \cite{wei2022chain}, which facilitates iterative reasoning over previously processed tokens.

In this work, we introduce a novel approach called \textbf{Masked Attention by Segment (MAS)}, which leverages pre-trained GPT models and adjust them to utilize information from future tokens.
MAS is inspired by the concept of unified architecture but eliminates the need for training new architectures from scratch.
{MAS achieves this by modifying the attention masking mechanism of GPTs, unmasking entire segments of input prompts. These segments appear as square blocks in Fig.~\ref{fig:teaser}.
During autoregressive processing, the model reverts to using causal attention, selectively unmasking the attention between future and earlier tokens that are not available outside of training (the diagonal segments in the figure).}

On a set of commonsense reasoning tasks, we show how our MAS lightweight adaption can push publicly available GPTs to new state-of-the-art (SOTA) results, as summarized in Fig.~\ref{fig:summaryresults}.

\section{Related Work}

Encoder-only models like BERT \cite{devlin2018bert} are primarily designed for bidirectional understanding tasks and excel in applications such as classification and question answering. While BERT can generate text autoregressively, each newly generated token changes the attention computation, requiring all dependent hidden states to be recomputed. Unlike decoder-only models, which use a key-value (KV) cache to efficiently generate multiple tokens during inference, BERT’s bidirectional design prevents such optimization. This makes BERT impractical for token-by-token decoding.

The T5 framework \cite{raffel2020exploring}, with its encoder-decoder architecture, is effective for many NLP tasks due to its ability to incorporate bidirectional context during encoding and causal generation during decoding.
However, SOTA and efficient performances in text generation are dominated by decoder-only models with causal masking, such as GPT-based architectures \cite{brown2020language,jiang2023mistral,yang2024qwen2,dubey2024llama,abdin2024phi4technicalreport}.
These models are highly efficient for token-by-token generation but cannot fully utilize input prompt information in their current design.

The most closely related work to our approach, PrefixLM, was explored in the T5 framework \cite{raffel2020exploring}. PrefixLM operates within a unified decoder-only architecture but enables bidirectional attention over a designated prefix of the input sequence while maintaining causal attention for the remainder.
However, PrefixLM requires training from scratch and is limited to single-turn inputs, overlooking scenarios with multiple prefill phases, as often encountered in chat-based systems.

In contrast, our approach enables the easy enhancement of SOTA decoder-only models by unlocking the potential of bidirectional attention in non-generated segments through lightweight fine-tuning.
Trained on massive corpora with causal masking, these models can be enhanced with limited hardware and just a few hours of fine-tuning, enabling them to effectively use bidirectional attention during the prefill phase.

Furthermore, MAS is specifically designed for chat-based tasks, where input prompts are naturally segmented into components such as system instructions and user queries.
Unlike PrefixLM, which processes the input as a single block, MAS applies bidirectional attention within each segment while preserving causal masking for generated outputs.
This segmentation-aware approach enables caching of the system prompt's key-value (KV) cache, as the system prompt typically remains unchanged across different user sessions. This reduces redundant computations and significantly lowers the latency for generating the first token.

\section{Background}
\label{Background}

Generative Pre-trained Transformer (GPT) models, also known as decoder-only transformers, are a family of language models (LMs) that generate each new token based on all previously produced tokens. The input, referred to as a \textit{prompt}, is a sequence of \( n \) tokens denoted as \([t^{1}, \ldots, t^{n}]\).

Processing begins with a \textbf{prefill} phase, in which the model consumes the entire prompt at once to predict the next token \( t^{n+1} \). Afterwards, the model enters an \textbf{autoregressive decoding} phase, incrementally generating subsequent tokens \( t^{n+k} \) for \( k = 2, 3, \ldots \). At the \( k \)-th step, the input is the concatenation of the original prompt and all previously generated tokens:
\begin{equation}
[t^{1}, \ldots, t^{n}, t^{n+1}, \ldots, t^{n+k-1}].
\end{equation}

This autoregressive process continues until the model reaches a predefined generation limit or produces a designated stop token. The resulting sequence,
\begin{equation}
[t^{1}, \ldots, t^{n+k}],
\end{equation}
can then serve as a new prompt, initiating another cycle of prefill and decoding if needed.

\subsection{Architecture}
\label{Architecture}
GPT architectures are built on a residual stream that connects multiple transformer blocks, each comprising three main components: a multi-head attention block (\textbf{Attn}), a multi-layer perceptron block (\textbf{MLP}), and layer normalization (\textbf{LN}) applied before each block.

The attention block operates using four parameter matrices: the query matrix \(W_q \in \mathbb{R}^{d \times d}\), key matrix \(W_k \in \mathbb{R}^{d \times d}\), value matrix \(W_v \in \mathbb{R}^{d \times d}\), and output matrix \(W_o \in \mathbb{R}^{d \times d}\). These matrices are divided into \(h\) heads, with each head using partitioned matrices of dimensions \(\mathbb{R}^{d \times \frac{d}{h}}\). For the \(\ell\)-th head, the projections are given as:
\begin{align}
    Q^\ell &= X W_q^\ell, \\
    K^\ell &= X W_k^\ell, \\
    V^\ell &= X W_v^\ell,
\end{align}
where \(Q^\ell, K^\ell, V^\ell \in \mathbb{R}^{n \times \frac{d}{h}}\) and \(X = [x_1 \dots x_n] \in \mathbb{R}^{n \times d}\) represents the input sequence.

The attention scores for each head are computed using the position-encoded query and key by:
\begin{equation}
\label{eq:attn}
A^\ell = \text{softmax}\left(\frac{\tilde{Q}^\ell \tilde{K}^{\ell \top}}{\sqrt{d/h}} + M\right),
\end{equation}
where \(\tilde{Q}^\ell\) and \(\tilde{K}^\ell\) are the query and key matrices after applying Rotary Position Embedding (RoPE)~\cite{su2024roformer}.

Here, \(M \in \mathbb{R}^{n \times n}\) enforces causal masking, ensuring that each token attends only to tokens that precede it in the sequence. This constraint is critical for maintaining the autoregressive property of GPT models, where predictions are conditioned only on prior tokens.

Each attention head output is computed as \(A^\ell V^\ell\), and the outputs from all heads are concatenated and projected using \(W_o\):
\begin{equation}
    \text{Attn}(X) = [A^1 V^1, \ldots, A^h V^h] W_o.
\end{equation}

Each attention layer is followed by an MLP layer. With the SwiGLU variant \cite{shazeer2020glu}, the MLP architecture in the examined model uses three weight matrices: \( W_U \), \( W_G \), \( W_D^\top \in \mathbb{R}^{d \times d_m}\), 
 along with an activation function such as SiLU, \( f \).
 The MLP output is computed as:
\begin{equation}
    \text{MLP}(X) = (f(X W_U) \odot (X W_G)) W_D
\end{equation}

The output of the \(i\)-th transformer block, including layer normalization (LN), is computed as:
\begin{align}
    X^i_{\text{Attn}} &= X^i + \text{Attn}(\text{LN}(X^i)), \\
    X^{i+1} &= X^i_{\text{Attn}} + \text{MLP}(\text{LN}({X}^i_{\text{Attn}})).
\end{align}
The attention and MLP blocks complement each other: the attention mechanism captures dependencies between tokens across the sequence, while the MLP refines these representations independently at each token. Together, they iteratively enhance the hidden representation as it flows through the transformer layers, with layer normalization stabilizing the output at each step.

\subsection{Fine Tuning}
\label{Fine Tuning}
In this work, we examine publicly available GPTs from HuggingFace \cite{wolf2019huggingface}, including Llama \cite{touvron2023llama}, Qwen \cite{yang2024qwen2}, and Mistral \cite{jiang2023mistral}. These models are pre-trained on extensive corpora to support general-purpose applications.  

A common approach to enhance a GPT performance for specific tasks, such as reasoning or conversational applications, involves fine-tuning on task-specific datasets. However, full fine-tuning is computationally expensive. To address this challenge, adapter-based fine-tuning methods \cite{peft} have been developed, offering a computationally efficient alternative while maintaining comparable performance.  
Specifically, we focus on fine-tuning these general-purpose models using LoRA \cite{hu2022lora}, the most widely adopted adapter-based fine-tuning technique.

LoRA (Low-Rank Adaptation) fine-tuning modifies only the \textbf{weight matrices}.
Specifically, the original LoRA paper suggests to modify the attention mechanism only by fine-tuning the query matrix \(W_q\) and the value matrix \(W_v\).
Instead of directly updating the full matrices \(W_q\) and \(W_v\), LoRA introduces low-rank decomposition matrices \(A\) and \(B\) such that the updated matrices become:

\begin{align}
    \tilde{W}_q &= W_q + \Delta W_q = W_q + A_q B_q, \\
    \tilde{W}_v &= W_v + \Delta W_v = W_v + A_v B_v,
\end{align}

where \(A_q, B_q^\top \in \mathbb{R}^{d \times r}\) and \(A_v, B_v^\top \in \mathbb{R}^{d \times r}\), with \(r \ll d\) being the rank of the decomposition. These low-rank matrices are trained during fine-tuning, while the original weights \(W_q\) and \(W_v\) remain frozen. This approach reduces the number of trainable parameters significantly, making fine-tuning more efficient.

By modifying only the query and value matrices with low-rank updates, LoRA enables efficient fine-tuning while preserving the pre-trained knowledge stored in the original model weights.

\section{Method}
\label{Method}

Causal attention, as an autoregressive mechanism, restricts information flow to propagate only from earlier tokens to later ones. While this design is essential during the autoregressive decoding phase, it is unnecessarily restrictive during the prefill phase, in which the entire prompt is available at once. Specifically, causal masking prevents the model from leveraging information from later tokens in the prompt, introducing a constraint in the attention computation.

For instance, consider the following example from the commonsense reasoning task ARC-Challenge \cite{clark2018think}:
\textit{Please choose the correct answer to the question: Giant sloths lived in the Americas and the Caribbean during the Ice Age... Most of these sloths disappeared... some of these sloths lived alongside humans. What is the most likely reason that these last giant sloths became extinct?  
Answer1: disease... Answer4: humans as predators...''}

To correctly answer this question, the model must infer a specific information from the prompt (that human lived next to giant sloths) and ignore other (the fact that sloths lived through the Ice Age is a distracting detail).
In a standard autoregressive model with causal masking, the text is read unidirectionally. This means the model cannot contextualize the final question while reading the initial sentences. Its success relies solely on its ability to memorize the prompt, as it cannot revisit earlier parts of the text when processing subsequent information.
In contrast, humans readers can revisit earlier sentences or questions to focus on relevant details and build a coherent understanding.

The unidirectional nature of GPTs with causal masking imposes a limitation on their capabilities to integrate context from the entire prompt.
To address this limitation, we propose {Masked Attention by Segment (MAS)}, which adapts the attention mechanism to process full prompts more effectively. Similar to encoder-decoder, MAS operates in two modes:

\noindent{\bf i. Prefill Phase:\quad} MAS removes the strict causal masking within each input prompt, allowing tokens to attend to both earlier and later tokens in the same block.

\noindent{\bf ii. Autoregressive Phase:\quad} MAS reverts to standard causal masking, which ensures that each token only attends to preceding tokens. During training, this masking reflects the autoregressive nature of text generation, where future tokens are inaccessible at inference time.

Unlike previous approaches, MAS is designed for use with pre-trained GPT models, eliminating the need to train new models from scratch. To apply MAS, we modify the attention masking mechanism within general-purpose GPTs, and finetune the model on a dataset according to its downstream task.
Specifically, the mask $M$ in Eq.~\ref{eq:attn} is defined in the conventional GPT models as:
\begin{equation}
M_{i, j} = 
\begin{cases} 
0 & \text{if } i \leq j, \\
-\infty & \text{if } i > j.
\end{cases}
\end{equation}

where \( M \in \mathbb{R}^{n \times n} \) and \( i, j \in \{1, 2, \ldots, n\} \) are token indices.

In MAS, the mask becomes
\begin{equation}
M_{i, j} = 
\begin{cases} 
0 & \text{if } i \leq j \text{ or } S(i) = S(j), \\
-\infty & \text{otherwise}.
\end{cases}
\end{equation}

\noindent where \( S(i) \) is a function that returns the segment ID of the token at position \( i \). Tokens within the same prefill segment (such as a system prompt or user prompt) share the same segment ID.

By restricting modifications solely to the attention mask, MAS maintains computational efficiency and seamlessly integrates with existing architectures.

In chat-based tasks, MAS leverages the structured nature of interactions, treating inputs as distinct blocks: the \textbf{system prompt} (providing initial instructions or context) and the \textbf{user prompt} (the user’s actual input). These blocks are unmasked during the prefill phase, enabling tokens within the same block to attend to both earlier and later tokens. In contrast, the \textbf{assistant tokens} (model-generated response) use standard causal masking, ensuring that each token only attends to previously generated tokens.

During the training phase, given a set of chat-template examples, the prefill blocks are identified by special tokens that mark the beginning of the system and the user prompt segments. During test-time inference, MAS identifies the prompts as the inputs. When the model generates its response, it switches to causal masking.

A key design choice of MAS is the separation between the system prompt block and subsequent user prompt blocks.
Today, commercial GPTs like ChatGPT operate with a substantially long and fixed system prompts that provide essential context. Instead of processing this system prompt with each user input, it can be preprocessed and cached as a numerical representation, reducing computational redundancy.
The separation between system and user prompts in MAS enables this caching functionality.

\begin{table*}
\begin{small}
\begin{tabular}{lccccccccr}
\toprule
Model & BoolQ & PIQA & SIQA & HellaSwag & WinoGrande & ARC-e & ARC-c & OBQA & Avg. \\
\midrule
\midrule
GPT 3.5-turbo CoT & 73.1 & 85.4 & 68.5 & 78.5 & 66.1 & 89.8 & 79.9 & 74.8 & 77.0 \\
\midrule
Llama-3-8B & 74.5 & 87.8 & 79.2 & 94.5 & 84.7 & 89.4 & 77.2 & 84.4 & 84.0 \\
Llama-3-8B+MAS & 74.9 & 88.6 & 81.5 & 93.2 & 88.0 & 92.3 & 81.5 & 86.8 & \textbf{85.8} \\
\midrule
Llama-3.1-8B & 73.5 & 87.5 & 80.3 & 94.4 & 85.3 & 88.8 & 78.0 & 84.8 & 84.1 \\
Llama-3.1-8B+MAS & 74.8 & 88.0 & 82.0 & 91.6 & 86.0 & 92.2 & 81.7 & 85.8 & \textbf{85.2} \\
\midrule
Llama-3.2-1B & 62.1 & 72.1 & 71.0 & 58.7 & 65.4 & 67.0 & 49.3 & 61.0 & 63.3 \\
Llama-3.2-1B+MAS & 62.4 & 78.2 & 73.0 & 69.1 & 70.2 & 72.7 & 55.5 & 67.6 & \textbf{68.6} \\
\midrule
Llama-3.2-3B & 70.0 & 83.4 & 77.2 & 90.7 & 79.5 & 83.0 & 70.6 & 78.0 & 79.0 \\
Llama-3.2-3B+MAS & 71.1 & 86.0 & 79.6 & 90.6 & 83.9 & 89.4 & 75.1 & 82.6 & \textbf{82.3} \\
\midrule
Mistral-7B-v0.1 & 74.3 & 88.4 & 80.0 & 94.8 & 85.6 & 88.8 & 78.8 & 86.6 & 84.7 \\
Mistral-7B-v0.1+MAS & 70.9 & 88.5 & 82.5 & 91.4 & 88.2 & 92.6 & 80.5 & 89.2 & \textbf{85.5} \\
\midrule
Qwen2.5-3B & 60.4 & 84.7 & 75.9 & 66.7 & 77.4 & 92.8 & 81.7 & 85.0 & 78.1 \\
Qwen2.5-3B+MAS & 68.3 & 85.6 & 80.6 & 90.7 & 83.8 & 93.5 & 84.0 & 86.2 & \textbf{84.1} \\
\midrule
Qwen2.5-7B & 72.2 & 90.1 & 79.4 & 94.4 & 83.3 & 95.2 & 87.6 & 90.6 & 86.6 \\
Qwen2.5-7B+MAS & 73.7 & 90.3 & 83.5 & 94.5 & 88.8 & 96.7 & 90.1 & 93.0 & \textbf{88.8} \\
\midrule
\midrule
MAS win rate & 85.7 & 100.0 & 100.0 & 42.9 & 100.0 & 100.0 & 100.0 & 100.0 & 100.0 \\
\midrule
\bottomrule
\end{tabular}
\end{small}
\vspace{-2pt}
\caption{Results on the Commonsense Reasoning benchmark.}
\label{commonsense main results}
%\end{table*}
\smallskip
\smallskip
%\begin{table*}
\begin{small}
\begin{tabular}{lcccccccccr}
\toprule
model & train $\xrightarrow{}$ test & BoolQ & PIQA & SIQA & HellaSwag & WinoGrande & ARC-e & ARC-c & OBQA & Avg. \\
\midrule
\midrule

 \multirow{2}{1em}{1B} & MAS $\xrightarrow{}$ CA & 62.2 & 50.3 & 31.7 & 23.4 & 52.8 & 18.6 & 22.4 & 20.0 & 35.1 \\
 % & MAS $\xrightarrow{}$ MAS & 62.4 & 78.2 & 73.0 & 69.1 & 70.2 & 72.7 & 55.5 & 67.6 & 68.6 \\
 % & CA $\xrightarrow{}$ CA & 62.1 & 72.1 & 71.0 & 58.7 & 65.4 & 67.0 & 49.3 & 61.0 & 63.3 \\
 & CA $\xrightarrow{}$ MAS & 0.0 & 0.0 & 0.1 & 0 & 0.0 & 0.0 & 0.0 & 0.0 & 0.0 \\
\midrule
\multirow{2}{1em}{3B} & MAS $\xrightarrow{}$ CA & 62.4 & 78.2 & 68.8 & 22.6 & 71.0 & 81.1 & 63.3 & 68.6 & 64.4 \\
  % & MAS $\xrightarrow{}$ MAS & 71.1 & 86.0 & 79.6 & 90.6 & 83.9 & 89.4 & 75.1 & 82.6 & 82.3 \\
  % & CA $\xrightarrow{}$ CA & 70.0 & 83.4 & 77.2 & 90.7 & 79.5 & 83.0 & 70.6 & 78.0 & 79.0 \\
  & CA $\xrightarrow{}$ MAS & 19.6 & 2.4 & 5.9 & 0.1 & 0.9 & 5.3 & 4.7 & 5.6 & 5.5 \\
\bottomrule
\end{tabular}
\end{small}
\caption{Average accuracy on the Commonsense Reasoning tasks for different attention mechanisms while training and evaluating Llama-3.2 .
CA stands for causal attention.
{The results for MAS $\xrightarrow{}$ MAS and CA $\xrightarrow{}$ CA can be found in Tab.~\ref{commonsense main results} above.}
}
\label{tab: ablation mechanisem}
\end{table*}

\begin{figure*}[t]
\centering
\centerline{\includegraphics[width=0.68\linewidth]{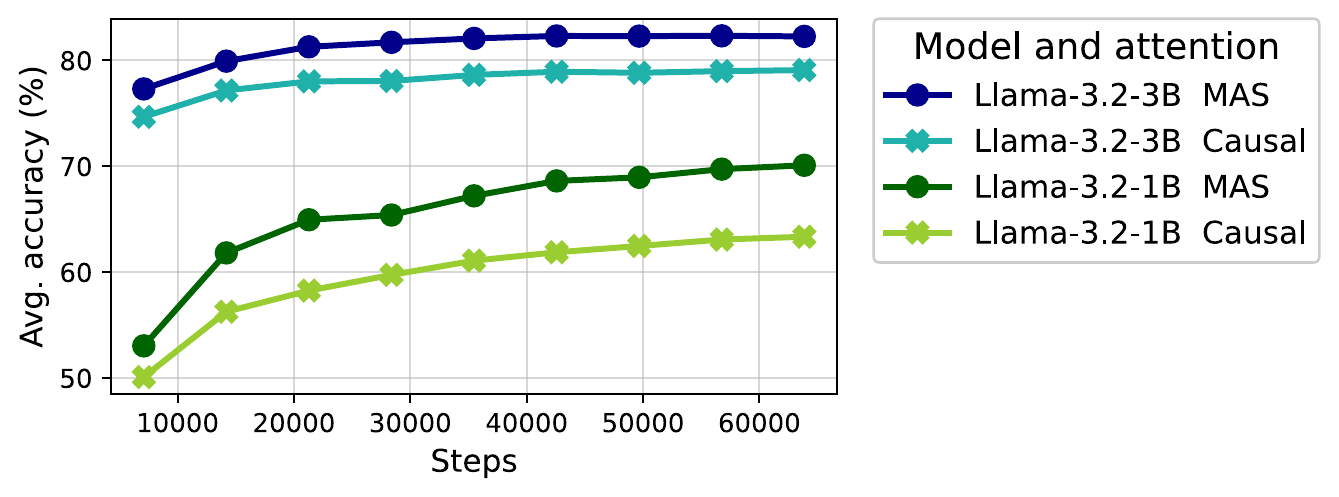}}
\vspace{-4pt}
\caption{The average accuracy on the Commonsense Reasoning during the fine-tuning of Llama-3.2, either 1B or 3B using conventional (causal) attention or MAS.}
\label{fig: training steps}
\end{figure*}

\begin{figure*}[t]
\centering
  \includegraphics[width=0.80\linewidth]{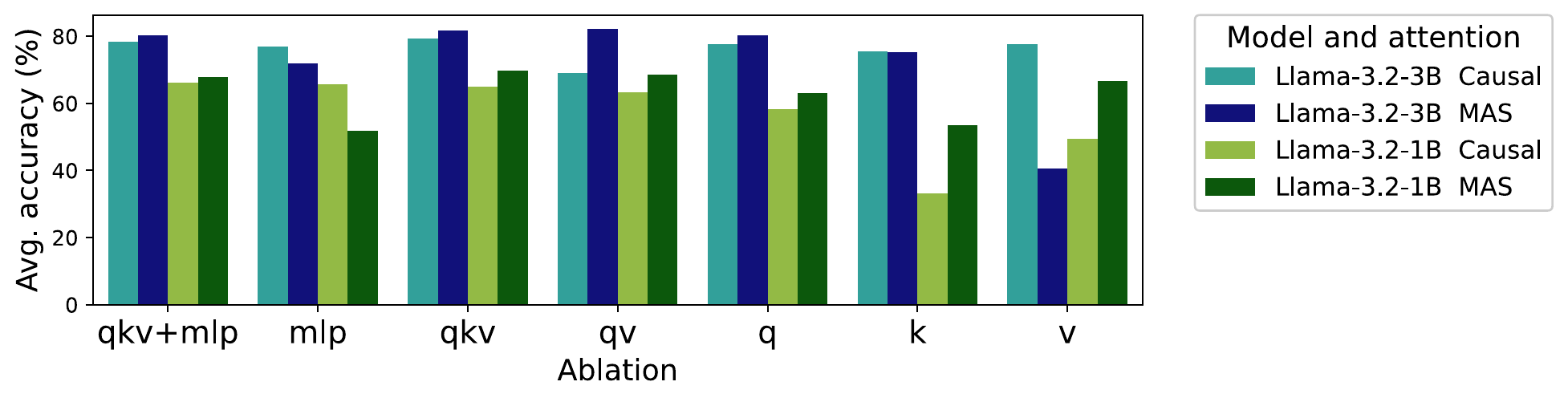}
  \vspace{-4pt}
  \caption{The average accuracy on the Commonsense Reasoning when training different matrices of Llama-3.2.}
    \label{fig: commonsense results QKVUD}
\end{figure*}
\begin{figure*}[t]
\begin{center}
\centerline{\includegraphics[width=\linewidth]{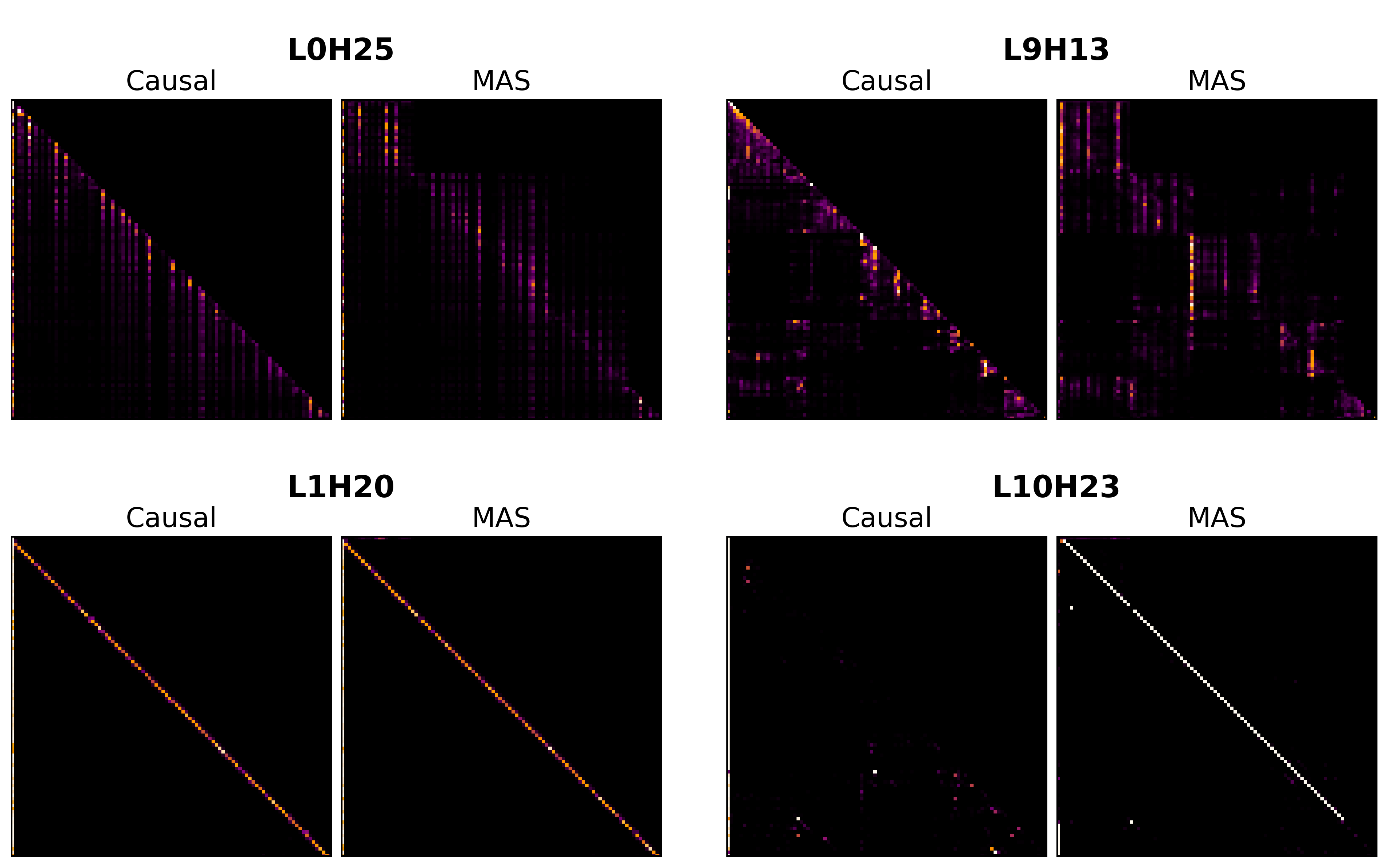}}
\caption{
The attention maps of two instances of Llama-3.2-1B fine-tuned for commonsense reasoning tasks, one with standard causal masking and the other with MAS, reveal distinct patterns: (i) L0H25 (layer 0, head 25) exhibits \textbf{N-gram Patterns}, indicating attention focused on sequences of consecutive tokens. (ii) L9H13 demonstrates \textbf{Block-Specific Patterns}, concentrated within defined blocks of the prompts. (iii) L10H23 showcases a \textbf{Forward-Looking} behavior, attending precisely to the next tokens within the same block. (iv) L1H20 serves as an example of \textbf{Preserved Patterns}.}
\label{fig: MAS viz main}
\end{center}
\vspace{-6pt}
\end{figure*}

\section{Experiments}
\label{Experiments}
\label{Commonsense Reasoning}
We evaluate MAS as an alternative to standard causal masking. In the following experiments, we fine-tune a set of GPT models, once using naive causal masking and once using MAS. To minimize computational costs, all fine-tuning setups are conducted with LoRA.

Our evaluation follows the experimental setup described in \citet{dora} and \citet{hu2023llm}, where models are fine-tuned on a dataset containing 170,000 samples of reasoning tasks in the template of chat-based prompts.
{The training and test samples are from eight distinct datasets, each designed to evaluate different reasoning abilities.
Some datasets, like BoolQ \cite{clark2019boolq}, provide naturally occurring yes/no questions about general knowledge, such as ``Is your body temperature lower in the morning?''.
Others, such as WinoGrande \cite{ai2:winogrande}, are fill-in-the-blank tasks with prompts like ``It was easy for Amy but not Rachel to create a meal because [blank] had taken woodshop in school. Option 1: Amy, Option 2: Rachel''.
While couple datasets evaluate scientific reasoning, such as the ARC Challenge \cite{clark2018think}, others, like HellaSwag \cite{zellers2019hellaswag}, focus on narrative completion, and PIQA \cite{Bisk2020} on physical commonsense reasoning.}

All samples are structured as multiple-choice question-answering tasks and are written in an instruction-based format. The system prompt states: ``Below is an instruction... Write a response that appropriately completes the request'', followed by the user prompt, presenting the actual question, such as ``Please answer the following question...''.

This diverse range of tasks enables us to assess model performance across various reasoning abilities, offering a comprehensive evaluation of MAS's effectiveness in enhancing GPT-based models.

The hyperparameters used for all models are detailed in \autoref{Hyperparameters}.
These settings are primarily based on the implementation used by \citet{dora}, with the exception of the batch size, which was adjusted due to computational constraints, and therefore also the learning rate.  
To ensure a fair evaluation, for each model we explored several learning rates in the range of $[1e-5,1e-3]$.
The learning rates selected for both masking approaches are the ones that achieved the highest average accuracy with causal masking, i.e., performance might be higher for MAS with other learning rates.

Our experiments include models from the Llama family \cite{dubey2024llama}, Qwen 2.5 \cite{qwen2.5}, and Mistral-7B-v0.1 \cite{jiang2023mistral}.
For comparison, we also include results for GPT 3.5-turbo \cite{radfordimproving} when using a zero shot chain-of-thoughts (CoT), as provided by \citet{hu2023llm}.

The results in Table~\ref{commonsense main results} illustrate that Masked Attention by Segment (MAS) consistently outperforms causal masking across most tasks and models.  
While the improvements are modest for some tasks and models, they are still significant: for example, MAS achieves an average accuracy increase of approximately $1.2\%$ on BoolQ and PIQA.
For the remaining tasks, the improvements range from $3.5\%$ to $4.75\%$. These gains translate to an overall increase in average accuracy of around $1\%$ to $7\%$ across different models. 
When analyzing performance on individual tasks, MAS delivers better results for all models in nearly every case, as reflected in the MAS win rate shown in the last row of the table. The only exceptions are a single model that did not improve on BoolQ and the HellaSwag task, where MAS underperforms.

The drop in performance on HellaSwag may be attributed to the experimental setup, which follows previous work \cite{dora} and employs a maximum sequence length cutoff during fine-tuning. HellaSwag is unique among the tasks due to its longer prompt lengths, implying that such a configuration may not allow MAS to fully adapt to the demands of tasks requiring extensive context.

Overall, these findings suggest that MAS has the potential to enhance the performance of downstream tasks of pretrained LLMs, which were originally trained using causal masking. A natural question is whether adapting to a new attention scheme requires more elaborate training. 
In our experiments, models were trained for 3 epochs (63,000 steps), with checkpoints saved every one-third of an epoch (7,000 steps). Using Llama-3.2 as a case study, we present the average accuracy at each checkpoint in \autoref{fig: training steps}. The results demonstrate that MAS consistently outperforms causal masking at every stage of training.

During the initial epoch, the accuracy gap between MAS and causal masking grows steadily, with MAS showing incremental improvements over causal masking. After the first epoch, performance gains slow for both methods, as training approaches a plateau. However, the performance gap established during the early stages remains consistent throughout the remaining training steps. 

These findings highlight two key points: i. Rapid adaptation: MAS achieves significant performance improvements early in training, and ii. Sustained progress: MAS maintains a stable advantage throughout training, with consistent improvement and no sharp fluctuations between steps.

{Next we examine whether the fine-tuning is essential for MAS, and whether models fine-tuned with MAS can perform well using only causal masking.
The results, shown in \autoref{tab: ablation mechanisem}, reveal that models fine-tuned with causal masking but evaluated with MAS perform poorly, emphasizing the necessity of fine-tuning for adapting to MAS.  
In contrast, models fine-tuned with MAS but evaluated with causal masking achieve moderate performance, with accuracy approximately half of what they achieve under MAS evaluation.}

{Another ablation study explored the effect of training different subsets of models' parameters, specifically the attention matrices, $W_q, W_k, W_v$, and the MLP matrices, $W_U, W_D$. Results, summarized in \autoref{fig: commonsense results QKVUD}, highlight:} \noindent{(i):} Training only the $W_q, W_v$ matrices achieves the best performance, with Llama-3.2-3B reaching $82.3\%$ accuracy, compared to $81.7\%$ for $W_q, W_k, W_v$ and $80.4\%$ for $W_q, W_k, W_v, W_U, W_D$. \noindent{(ii):} Surprisingly, training only the $W_q$ matrix achieves $80.3\%$ accuracy, underscoring the critical role of $W_q$ in positional embeddings via the RoPE mechanism. \noindent{(iii):} Across all setups, MAS consistently outperforms causal masking, except when training only the MLP matrices ($W_U, W_D$).
This aligns with the fact that MAS modifies the attention mechanism, making the fine-tuning of the attention matrix essential for leveraging MAS benefits.

For detailed results on the application of MAS when fine-tuning the MLP matrices, refer to \autoref{Additional Commonsense Reasoning Results}.  
Additionally, \autoref{Ablation Study and Training Analysis} presents an in-depth ablation study on the hyperparameters involved in the fine-tuning process.

\label{Interpretation}
To gain insights into how MAS influences model behavior, we \textbf{visualize attention maps} of models fine-tuned with MAS. In models trained with standard causal masking, attention maps exhibit a lower triangular structure, reflecting the restriction that each token can only attend to preceding tokens.
In contrast, MAS introduces a block-based structure in the attention maps, allowing tokens within the same block to attend to each other. 

\autoref{fig: MAS viz main} provides a comparison between two instances of Llama-3.2-1B, as detailed in \autoref{Commonsense Reasoning}: one fine-tuned with standard causal masking and the other fine-tuned with MAS. The input prompt in this comparison consists of three distinct components: the {system prompt}, the {user prompt}, and the {assistant's output}.
In our analysis, we identified four distinct attention patterns emerging across MAS attention heads:
\begin{itemize}[left=-10pt..0pt, itemindent=!,itemsep=-5pt,topsep=0pt]
    \item \textbf{Preserved Patterns:} Certain heads maintain their original behavior from standard causal masking, attending primarily to preceding tokens as expected.
    \item \textbf{Block-Specific Patterns:} A frequently observed pattern in attention maps is the presence of vertical lines indicating specific tokens within the prompt to which many other tokens attend. In causal attention, these vertical lines are confined below the main diagonal, but in MAS, they extend above the diagonal or even span the height of an entire block. Occasionally, this pattern highlights the boundaries of a block by emphasizing its first or last token.
    \item \textbf{N-gram Patterns:}
    This pattern appears as short crossed vertical or horizontal lines over the main diagonal, typically, spanning only a few tokens. It suggests MAS's ability to identify the context of a specific token using its neighboring tokens, including those preceding and following it. While causal attention enables context identification from preceding tokens, MAS extends this capability to include future tokens as well.
    \item \textbf{Forward-Looking Patterns:} A Certain attention heads focus on tokens located exactly a few steps ahead, identifiable by cross secondary diagonals above the main diagonal.
\end{itemize}

These patterns are illustrated in \autoref{fig: MAS viz main}, with a particularly notable example of the forward-looking pattern: in layer 10, head 23, an almost perfect diagonal line appears just above the main diagonal of the attention matrix.
This behavior demonstrates the ability of certain heads to anticipate and focus on subsequent tokens within the same block. Importantly, this pattern aligns with MAS’s design, as it breaks between the system and user prompts, respecting the separation of blocks. This forward-looking attention terminates at the start of the assistant's output, where causal masking resumes during the autoregressive phase to ensure proper token-by-token generation. Additional results are provided in \autoref{MAS visualization}.

\section{Conclusions}
\label{Conclusions}
In this work, we introduced Masked Attention by Segment (MAS), a novel method to enhance the performance of GPT-based models by enabling them to leverage future tokens during input processing.
This approach addresses the inherent limitations of standard causal attention through a straightforward fine-tuning process.
Extensive experiments on commonsense reasoning benchmarks demonstrated the scalability and effectiveness of MAS across diverse tasks and training setups.
Notably, MAS consistently outperformed causal masking in accuracy, with improvements evident early in training and maintaining a stable advantage as training progressed.

\section*{Limitations}
While Masked Attention by Segment (MAS) offers a promising enhancement for GPT-based models, several limitations warrant consideration.  

First, MAS requires fine-tuning existing models.
Although this process is significantly less resource-intensive than training models from scratch, it still imposes computational and time constraints, which may limit its accessibility for broader adoption.  

Second, MAS shows reduced performance on tasks with exceptionally long prompts, such as HellaSwag.
We hypothesize that sequence length cutoffs during fine-tuning may prevent the model from fully leveraging MAS in these scenarios.
This suggests the need for further optimization or task-specific adaptations to improve its effectiveness on length-sensitive tasks.  

For evaluation, we selected models from today's most capable and widespread publicly available GPTs, such as Llama-3.2 and Qwen2.5, focusing on their base versions rather than their instruction-tuned counterparts.
This choice aligns with previous work \cite{dora}, which used base models to avoid the complexities introduced by model-specific templated prompts.
While this approach provides a consistent evaluation framework, we acknowledge that MAS may behave differently on other models or specialized setups.  

{We focused on commonsense reasoning benchmarks, drawn from prior studies, as a practical proxy for evaluating MAS's effectiveness. This approach, shaped by computational constraints, enabled a focused and comprehensive assessment of its capabilities.} Future research should explore MAS's generalizability to a wider range of tasks and its potential when training from scratch.

\smallskip
\smallskip
\section*{Ethics Statement}
{This work aims to enhance language models by introducing a novel method to improve their performance through advancements in the attention mechanism.
We recognize the potential of such technologies and emphasize the importance of their responsible use.
While our contributions are intended to support the development of more aligned models, we stress the need of preventing misuse, such as generating harmful content.
Future research should focus on promoting applications that align with societal benefit.}

\section*{Acknowledgements}
This work was supported by the Tel Aviv University Center for AI and Data Science (TAD).
Y. R. thanks the Career Advancement Fellowship, Technion, for providing research support.
Y. R. and L. R. were funded by the European Union (ERC, SafetyBounds, 101163414). Views and opinions expressed are however those of the author(s) only and do not necessarily reflect those of the European Union or the European Research Council Executive Agency. Neither the European Union nor the granting authority can be held responsible for them.

\bibliography{custom}

\begin{thebibliography}{29}
\providecommand{\natexlab}[1]{#1}

\bibitem[{Abdin et~al.(2024)Abdin, Aneja, Behl, Bubeck, Eldan, Gunasekar, Harrison, Hewett, Javaheripi, Kauffmann, Lee, Lee, Li, Liu, Mendes, Nguyen, Price, de~Rosa, Saarikivi, Salim, Shah, Wang, Ward, Wu, Yu, Zhang, and Zhang}]{abdin2024phi4technicalreport}
Marah Abdin, Jyoti Aneja, Harkirat Behl, Sébastien Bubeck, Ronen Eldan, Suriya Gunasekar, Michael Harrison, Russell~J. Hewett, Mojan Javaheripi, Piero Kauffmann, James~R. Lee, Yin~Tat Lee, Yuanzhi Li, Weishung Liu, Caio C.~T. Mendes, Anh Nguyen, Eric Price, Gustavo de~Rosa, Olli Saarikivi, Adil Salim, Shital Shah, Xin Wang, Rachel Ward, Yue Wu, Dingli Yu, Cyril Zhang, and Yi~Zhang. 2024.
\newblock \href {https://arxiv.org/abs/2412.08905} {Phi-4 technical report}.
\newblock \emph{Preprint}, arXiv:2412.08905.

\bibitem[{Bisk et~al.(2020)Bisk, Zellers, Bras, Gao, and Choi}]{Bisk2020}
Yonatan Bisk, Rowan Zellers, Ronan~Le Bras, Jianfeng Gao, and Yejin Choi. 2020.
\newblock Piqa: Reasoning about physical commonsense in natural language.
\newblock In \emph{Thirty-Fourth AAAI Conference on Artificial Intelligence}.

\bibitem[{Brown et~al.(2020)Brown, Mann, Ryder, Subbiah, Kaplan, Dhariwal, Neelakantan, Shyam, Sastry, Askell et~al.}]{brown2020language}
Tom Brown, Benjamin Mann, Nick Ryder, Melanie Subbiah, Jared~D Kaplan, Prafulla Dhariwal, Arvind Neelakantan, Pranav Shyam, Girish Sastry, Amanda Askell, et~al. 2020.
\newblock Language models are few-shot learners.
\newblock \emph{Advances in neural information processing systems}, 33:1877--1901.

\bibitem[{Clark et~al.(2019)Clark, Lee, Chang, Kwiatkowski, Collins, and Toutanova}]{clark2019boolq}
Christopher Clark, Kenton Lee, Ming-Wei Chang, Tom Kwiatkowski, Michael Collins, and Kristina Toutanova. 2019.
\newblock Boolq: Exploring the surprising difficulty of natural yes/no questions.
\newblock In \emph{Proceedings of the 2019 Conference of the North American Chapter of the Association for Computational Linguistics: Human Language Technologies, Volume 1 (Long and Short Papers)}, pages 2924--2936.

\bibitem[{Clark et~al.(2018)Clark, Cowhey, Etzioni, Khot, Sabharwal, Schoenick, and Tafjord}]{clark2018think}
Peter Clark, Isaac Cowhey, Oren Etzioni, Tushar Khot, Ashish Sabharwal, Carissa Schoenick, and Oyvind Tafjord. 2018.
\newblock Think you have solved question answering? try arc, the ai2 reasoning challenge.
\newblock \emph{arXiv preprint arXiv:1803.05457}.

\bibitem[{Devlin(2018)}]{devlin2018bert}
Jacob Devlin. 2018.
\newblock Bert: Pre-training of deep bidirectional transformers for language understanding.
\newblock \emph{arXiv preprint arXiv:1810.04805}.

\bibitem[{Dong et~al.(2019)Dong, Yang, Wang, Wei, Liu, Wang, Gao, Zhou, and Hon}]{dong2019unified}
Li~Dong, Nan Yang, Wenhui Wang, Furu Wei, Xiaodong Liu, Yu~Wang, Jianfeng Gao, Ming Zhou, and Hsiao-Wuen Hon. 2019.
\newblock Unified language model pre-training for natural language understanding and generation.
\newblock \emph{Advances in neural information processing systems}, 32.

\bibitem[{Dubey et~al.(2024)Dubey, Jauhri, Pandey, Kadian, Al-Dahle, Letman, Mathur, Schelten, Yang, Fan et~al.}]{dubey2024llama}
Abhimanyu Dubey, Abhinav Jauhri, Abhinav Pandey, Abhishek Kadian, Ahmad Al-Dahle, Aiesha Letman, Akhil Mathur, Alan Schelten, Amy Yang, Angela Fan, et~al. 2024.
\newblock The llama 3 herd of models.
\newblock \emph{arXiv preprint arXiv:2407.21783}.

\bibitem[{Hu et~al.(2022)Hu, Shen, Wallis, Allen-Zhu, Li, Wang, Wang, and Chen}]{hu2022lora}
Edward~J Hu, Yelong Shen, Phillip Wallis, Zeyuan Allen-Zhu, Yuanzhi Li, Shean Wang, Lu~Wang, and Weizhu Chen. 2022.
\newblock \href {https://openreview.net/forum?id=nZeVKeeFYf9} {Lo{RA}: Low-rank adaptation of large language models}.
\newblock In \emph{International Conference on Learning Representations}.

\bibitem[{Hu et~al.(2023)Hu, Wang, Lan, Xu, Lim, Bing, Xu, Poria, and Lee}]{hu2023llm}
Zhiqiang Hu, Lei Wang, Yihuai Lan, Wanyu Xu, Ee-Peng Lim, Lidong Bing, Xing Xu, Soujanya Poria, and Roy Lee. 2023.
\newblock Llm-adapters: An adapter family for parameter-efficient fine-tuning of large language models.
\newblock In \emph{Proceedings of the 2023 Conference on Empirical Methods in Natural Language Processing}, pages 5254--5276.

\bibitem[{Jiang et~al.(2023)Jiang, Sablayrolles, Mensch, Bamford, Chaplot, Casas, Bressand, Lengyel, Lample, Saulnier et~al.}]{jiang2023mistral}
Albert~Q Jiang, Alexandre Sablayrolles, Arthur Mensch, Chris Bamford, Devendra~Singh Chaplot, Diego de~las Casas, Florian Bressand, Gianna Lengyel, Guillaume Lample, Lucile Saulnier, et~al. 2023.
\newblock Mistral 7b.
\newblock \emph{arXiv preprint arXiv:2310.06825}.

\bibitem[{Liu et~al.(2024)Liu, Wang, Yin, Molchanov, Wang, Cheng, and Chen}]{dora}
Shih-yang Liu, Chien-Yi Wang, Hongxu Yin, Pavlo Molchanov, Yu-Chiang~Frank Wang, Kwang-Ting Cheng, and Min-Hung Chen. 2024.
\newblock Dora: Weight-decomposed low-rank adaptation.
\newblock In \emph{Forty-first International Conference on Machine Learning}.

\bibitem[{Mangrulkar et~al.(2022)Mangrulkar, Gugger, Debut, Belkada, Paul, and Bossan}]{peft}
Sourab Mangrulkar, Sylvain Gugger, Lysandre Debut, Younes Belkada, Sayak Paul, and Benjamin Bossan. 2022.
\newblock Peft: State-of-the-art parameter-efficient fine-tuning methods.
\newblock \url{https://github.com/huggingface/peft}.

\bibitem[{Mihaylov et~al.(2018)Mihaylov, Clark, Khot, and Sabharwal}]{OpenBookQA2018}
Todor Mihaylov, Peter Clark, Tushar Khot, and Ashish Sabharwal. 2018.
\newblock Can a suit of armor conduct electricity? a new dataset for open book question answering.
\newblock In \emph{EMNLP}.

\bibitem[{Radford et~al.(2018)Radford, Narasimhan, Salimans, and Sutskever}]{radfordimproving}
Alec Radford, Karthik Narasimhan, Tim Salimans, and Ilya Sutskever. 2018.
\newblock Improving language understanding by generative pre-training.

\bibitem[{Raffel et~al.(2020)Raffel, Shazeer, Roberts, Lee, Narang, Matena, Zhou, Li, and Liu}]{raffel2020exploring}
Colin Raffel, Noam Shazeer, Adam Roberts, Katherine Lee, Sharan Narang, Michael Matena, Yanqi Zhou, Wei Li, and Peter~J Liu. 2020.
\newblock Exploring the limits of transfer learning with a unified text-to-text transformer.
\newblock \emph{Journal of machine learning research}, 21(140):1--67.

\bibitem[{Sakaguchi et~al.(2021)Sakaguchi, Bras, Bhagavatula, and Choi}]{ai2:winogrande}
Keisuke Sakaguchi, Ronan~Le Bras, Chandra Bhagavatula, and Yejin Choi. 2021.
\newblock Winogrande: An adversarial winograd schema challenge at scale.
\newblock \emph{Communications of the ACM}, 64(9):99--106.

\bibitem[{Sap et~al.(2019)Sap, Rashkin, Chen, Le~Bras, and Choi}]{siqa2019}
Maarten Sap, Hannah Rashkin, Derek Chen, Ronan Le~Bras, and Yejin Choi. 2019.
\newblock \href {https://doi.org/10.18653/v1/D19-1454} {Social {IQ}a: Commonsense reasoning about social interactions}.
\newblock In \emph{Proceedings of the 2019 Conference on Empirical Methods in Natural Language Processing and the 9th International Joint Conference on Natural Language Processing (EMNLP-IJCNLP)}, pages 4463--4473, Hong Kong, China. Association for Computational Linguistics.

\bibitem[{Shazeer(2020)}]{shazeer2020glu}
Noam Shazeer. 2020.
\newblock Glu variants improve transformer.
\newblock \emph{arXiv preprint arXiv:2002.05202}.

\bibitem[{Su et~al.(2024)Su, Ahmed, Lu, Pan, Bo, and Liu}]{su2024roformer}
Jianlin Su, Murtadha Ahmed, Yu~Lu, Shengfeng Pan, Wen Bo, and Yunfeng Liu. 2024.
\newblock Roformer: Enhanced transformer with rotary position embedding.
\newblock \emph{Neurocomputing}, 568:127063.

\bibitem[{Tay et~al.(2023)Tay, Dehghani, Tran, Garcia, Wei, Wang, Chung, Bahri, Schuster, Zheng, Zhou, Houlsby, and Metzler}]{tayul2}
Yi~Tay, Mostafa Dehghani, Vinh~Q. Tran, Xavier Garcia, Jason Wei, Xuezhi Wang, Hyung~Won Chung, Dara Bahri, Tal Schuster, Steven Zheng, Denny Zhou, Neil Houlsby, and Donald Metzler. 2023.
\newblock \href {https://openreview.net/forum?id=6ruVLB727MC} {{UL}2: Unifying language learning paradigms}.
\newblock In \emph{The Eleventh International Conference on Learning Representations}.

\bibitem[{Team(2024)}]{qwen2.5}
Qwen Team. 2024.
\newblock \href {https://qwenlm.github.io/blog/qwen2.5/} {Qwen2.5: A party of foundation models}.

\bibitem[{Touvron et~al.(2023)Touvron, Martin, Stone, Albert, Almahairi, Babaei, Bashlykov, Batra, Bhargava, Bhosale et~al.}]{touvron2023llama}
Hugo Touvron, Louis Martin, Kevin Stone, Peter Albert, Amjad Almahairi, Yasmine Babaei, Nikolay Bashlykov, Soumya Batra, Prajjwal Bhargava, Shruti Bhosale, et~al. 2023.
\newblock Llama 2: Open foundation and fine-tuned chat models.
\newblock \emph{arXiv preprint arXiv:2307.09288}.

\bibitem[{Vaswani et~al.(2017)Vaswani, Shazeer, Parmar, Uszkoreit, Jones, Gomez, Kaiser, and Polosukhin}]{vaswani2017attention}
Ashish Vaswani, Noam Shazeer, Niki Parmar, Jakob Uszkoreit, Llion Jones, Aidan~N Gomez, {\L}ukasz Kaiser, and Illia Polosukhin. 2017.
\newblock Attention is all you need.
\newblock \emph{Advances in neural information processing systems}, 30.

\bibitem[{Wei et~al.(2022)Wei, Wang, Schuurmans, Bosma, Xia, Chi, Le, Zhou et~al.}]{wei2022chain}
Jason Wei, Xuezhi Wang, Dale Schuurmans, Maarten Bosma, Fei Xia, Ed~H Chi, Quoc~V Le, Denny Zhou, et~al. 2022.
\newblock Chain-of-thought prompting elicits reasoning in large language models.
\newblock In \emph{Advances in Neural Information Processing Systems}.

\bibitem[{Wolf et~al.(2020)Wolf, Debut, Sanh, Chaumond, Delangue, Moi, Cistac, Rault, Louf, Funtowicz, Davison, Shleifer, von Platen, Ma, Jernite, Plu, Xu, Le~Scao, Gugger, Drame, Lhoest, and Rush}]{wolf2019huggingface}
Thomas Wolf, Lysandre Debut, Victor Sanh, Julien Chaumond, Clement Delangue, Anthony Moi, Pierric Cistac, Tim Rault, Remi Louf, Morgan Funtowicz, Joe Davison, Sam Shleifer, Patrick von Platen, Clara Ma, Yacine Jernite, Julien Plu, Canwen Xu, Teven Le~Scao, Sylvain Gugger, Mariama Drame, Quentin Lhoest, and Alexander Rush. 2020.
\newblock \href {https://doi.org/10.18653/v1/2020.emnlp-demos.6} {Transformers: State-of-the-art natural language processing}.
\newblock In \emph{Proceedings of the 2020 Conference on Empirical Methods in Natural Language Processing: System Demonstrations}, pages 38--45, Online. Association for Computational Linguistics.

\bibitem[{Yang et~al.(2024)Yang, Yang, Hui, Zheng, Yu, Zhou, Li, Li, Liu, Huang et~al.}]{yang2024qwen2}
An~Yang, Baosong Yang, Binyuan Hui, Bo~Zheng, Bowen Yu, Chang Zhou, Chengpeng Li, Chengyuan Li, Dayiheng Liu, Fei Huang, et~al. 2024.
\newblock Qwen2 technical report.
\newblock \emph{arXiv preprint arXiv:2407.10671}.

\bibitem[{Zellers et~al.(2019)Zellers, Holtzman, Bisk, Farhadi, and Choi}]{zellers2019hellaswag}
Rowan Zellers, Ari Holtzman, Yonatan Bisk, Ali Farhadi, and Yejin Choi. 2019.
\newblock Hellaswag: Can a machine really finish your sentence?
\newblock In \emph{Proceedings of the 57th Annual Meeting of the Association for Computational Linguistics}, pages 4791--4800.

\bibitem[{Zheng et~al.(2024)Zheng, Wang, Huang, Song, Yang, Tang, Xiong, and Li}]{zheng2024attention}
Zifan Zheng, Yezhaohui Wang, Yuxin Huang, Shichao Song, Mingchuan Yang, Bo~Tang, Feiyu Xiong, and Zhiyu Li. 2024.
\newblock Attention heads of large language models: A survey.
\newblock \emph{arXiv preprint arXiv:2409.03752}.

\end{thebibliography}

\appendix

% \newpage
% \clearpage

\begin{table*}[h!]
\begin{small}
\begin{tabular}{lccccccccr}
\toprule
model & BoolQ & PIQA & SIQA & HellaSwag & WinoGrande & ARC-e & ARC-c & OBQA & Avg. \\
\midrule
\midrule
Llama-3-8B & 75.1 & 88.6 & 81.0 & 95.2 & 87.1 & 90.7 & 78.8 & 84.4 & 85.1 \\
Llama-3-8B+MAS & 76.4 & 88.9 & 82.8 & 94.6 & 89.3 & 93.4 & 84.5 & 88.6 & \textbf{87.3} \\
\midrule
Llama-3.1-8B & 72.4 & 84.5 & 79.4 & 93.1 & 84.3 & 86.9 & 74.1 & 81.0 & 82.0 \\
Llama-3.1-8B+MAS & 70.6 & 85.4 & 80.3 & 90.6 & 84.1 & 88.8 & 76.5 & 84.6 & \textbf{82.6} \\
\midrule
Llama-3.2-1B & 62.8 & 74.1 & 72.8 & 65.7 & 67.7 & 68.6 & 52.3 & 65.2 & 66.2 \\
Llama-3.2-1B+MAS & 64.3 & 77.1 & 72.5 & 62.4 & 71.0 & 72.9 & 55.0 & 68.0 & \textbf{67.9} \\
\midrule
Llama-3.2-3B & 70.8 & 82.5 & 77.9 & 89.7 & 81.1 & 81.3 & 67.2 & 76.6 & 78.4 \\
Llama-3.2-3B+MAS & 70.0 & 84.4 & 79.6 & 87.4 & 80.4 & 86.7 & 72.5 & 82.0 & \textbf{80.4} \\
\midrule
Mistral-7B-v0.1 & 62.4 & 86.3 & 73.3 & 93.4 & 84.4 & 87.9 & 76.4 & 85.0 & 81.1 \\
Mistral-7B-v0.1+MAS & 72.2 & 86.2 & 79.0 & 82.9 & 85.0 & 90.1 & 77.2 & 83.0 & \textbf{82.0} \\
\midrule
Qwen2.5-3B & 70.2 & 85.3 & 78.7 & 91.8 & 80.3 & 91.9 & 79.4 & 84.8 & 82.8 \\
Qwen2.5-3B+MAS & 65.9 & 86.3 & 81.4 & 87.4 & 83.3 & 93.8 & 82.4 & 89.0 & \textbf{83.7} \\
\midrule
Qwen2.5-7B & 74.9 & 89.9 & 79.6 & 95.3 & 86.0 & 95.5 & 88.1 & 90.6 & 87.5 \\
Qwen2.5-7B+MAS & 76.1 & 90.7 & 82.9 & 94.1 & 87.5 & 96.7 & 89.6 & 92.6 & \textbf{88.8} \\
\midrule
\midrule
MAS win rate & 57.1 & 85.7 & 85.7 & 0.0 & 71.4 & 100.0 & 100.0 & 85.7 & 100.0 \\
\midrule
\bottomrule
\end{tabular}
\end{small}
\caption{Results of Commonsense Reasoning Tasks when training the $W_q, W_k, W_v$ matrices and the MLP's matrices $W_U, W_D$}
\label{tab: commonsense results QKVUD}
\end{table*}

\section{Additional Commonsense Reasoning Results}
\label{Additional Commonsense Reasoning Results}  
In \autoref{Commonsense Reasoning}, we presented results for the Commonsense Reasoning tasks. Here, we provide additional results and implementation details.

\subsection{Tasks}

The benchmarks and their implementations are based on \citet{hu2023llm}. The evaluation includes eight distinct tasks: BoolQ \cite{clark2019boolq}, PIQA \cite{Bisk2020}, SIQA \cite{siqa2019}, HellaSwag \cite{zellers2019hellaswag}, WinoGrande \cite{ai2:winogrande}, OBQA \cite{OpenBookQA2018}, ARC-e and ARC-c \cite{clark2018think}.
The fine-tuning dataset, Commonsense170K, consists of training samples from these benchmarks, formatted into structured templates that include a system prompt, additional input, and an answer for the model to complete.

\subsection{Trained Matrices}  
In \autoref{Commonsense Reasoning}, we focused on fine-tuning only two attention matrices ,$W_q,W_v$.
The original implementation of our experiments by \citet{dora} fine-tuned five matrices: the attention matrices $W_q, W_k, W_v$ and the MLP's $W_D, W_U$.
For completeness, we provide full results for the setup of fine-tuning the five matrices. 

The results in \autoref{tab: commonsense results QKVUD} reveal a pattern similar to fine-tuning only the $W_q$ and $W_v$ matrices, showing robust and consistent improvements in average performance across all models. However, as with the $W_q, W_v$ setup, MAS does not surpass causal masking on the HellaSwag benchmark. This limitation is likely due to the sequence length cutoff applied during fine-tuning, which constrains MAS’s ability to effectively process longer tasks like HellaSwag.

These results further support the idea that MAS is a promising approach for enhancing model performance.

\begin{table}[t]
\centering
\begin{tabular}{ll}
\hline
\textbf{Hyperparameters} & \textbf{Value} \\
\midrule
LoRA Rank & 32 \\ 
LoRA alpha & 64 \\ 
Dropout & 0.05 \\ 
Optimizer & AdamW \\ 
LR scheduler & Linear \\
Batch size & 8 \\ 
Cutoff length & 256 \\ 
Warmup steps & 100 \\ 
Epochs & 3 \\ 
\bottomrule
\end{tabular}
\caption{Common hyperparameters for Commonsense Reasoning fine-tuning}
\label{tab: hp commonsense}
\end{table}

\begin{table}[t]
\centering
\begin{tabular}{llll}
\hline
\textbf{Model} & \textbf{LR$_{qv}$} & \textbf{LR$_{Ex}$} & \textbf{Data type} \\
\midrule
Llama-3-8B & 2e-4 & 1e-4 & FP16 \\ 
Llama-3.1-8B & 2e-4 & 2e-4 & FP16 \\ 
Llama-3.2-1B & 2e-4 & 2e-4 & FP16 \\ 
Llama-3.2-3B & 8e-5 & 2e-4 & FP16 \\ 
Mistral-7B-v0.1  & 8e-5 & 3e-5 & FP16 \\ 
Qwen2.5-3B  & 1e-4 & 2e-4 &  BF16 \\ 
Qwen2.5-7B  & 1e-4 & 3e-4 & BF16 \\ 
\bottomrule
\end{tabular}
\caption{Model specific hyperparameters for Commonsense Reasoning fine-tuning}
\label{tab: lr commonsense}
\end{table}

\subsection{Hyperparameters}  
\label{Hyperparameters}  
The hyperparameters used for all models are detailed in \autoref{tab: hp commonsense}.
Model-specific hyperparameters are provided in \autoref{tab: lr commonsense}, where \textbf{LR}$_{qv}$ refers to the setup in which only the $W_q,W_v$ matrices are trained, and \textbf{LR}$_{Ex}$ corresponds to the configuration that includes training additional matrices ($W_q$, $W_k$, $W_v$, $W_U$, $W_D$).  
{For both causal and MAS we used the same hyperparameters.
We adjusted the learning rate between the two setups of \textbf{LR}$_{qv}$ and \textbf{LR}$_{Ex}$ after empirically we found the models did not reach full convergence results with the same learning rates.}

\paragraph{Computation}  
The experiments were conducted on machines equipped with Nvidia GPUs, including A5000, A6000, V-100, and A-100 models.

For example, in a LoRA fine-tuning setup with $10,000$ training steps, a batch size of 8 sentences, and a truncation cutoff at 256 tokens, the training process takes approximately 90 minutes for Llama-3.2-3B on an A5000 24GB GPU, for both MAS and causal attention. For Llama-3.2-1B, the same process takes around 40 minutes.
As we demonstrated, this duration is sufficient to adapt an existing GPT model to work effectively with MAS.

\section{Ablation Study and Training Analysis}
\label{Ablation Study and Training Analysis}
This section delves deeper into the fine-tuning process of GPTs using MAS, focusing on the Commonsense Reasoning task with Llama 3.2.  

\subsection{Hyperparameter Ablation Study}
To ensure the robustness of the results presented in \autoref{Commonsense Reasoning}, we conducted an ablation study across various hyperparameter configurations. Experiments were repeated with different learning rates and three random seeds. For efficiency, we limited training to 10,000 steps and evaluated on six tasks, excluding HellaSwag and BoolQ due to their higher computational costs.  

\begin{figure*}[h]
\centering
  \includegraphics[width=0.8\linewidth]{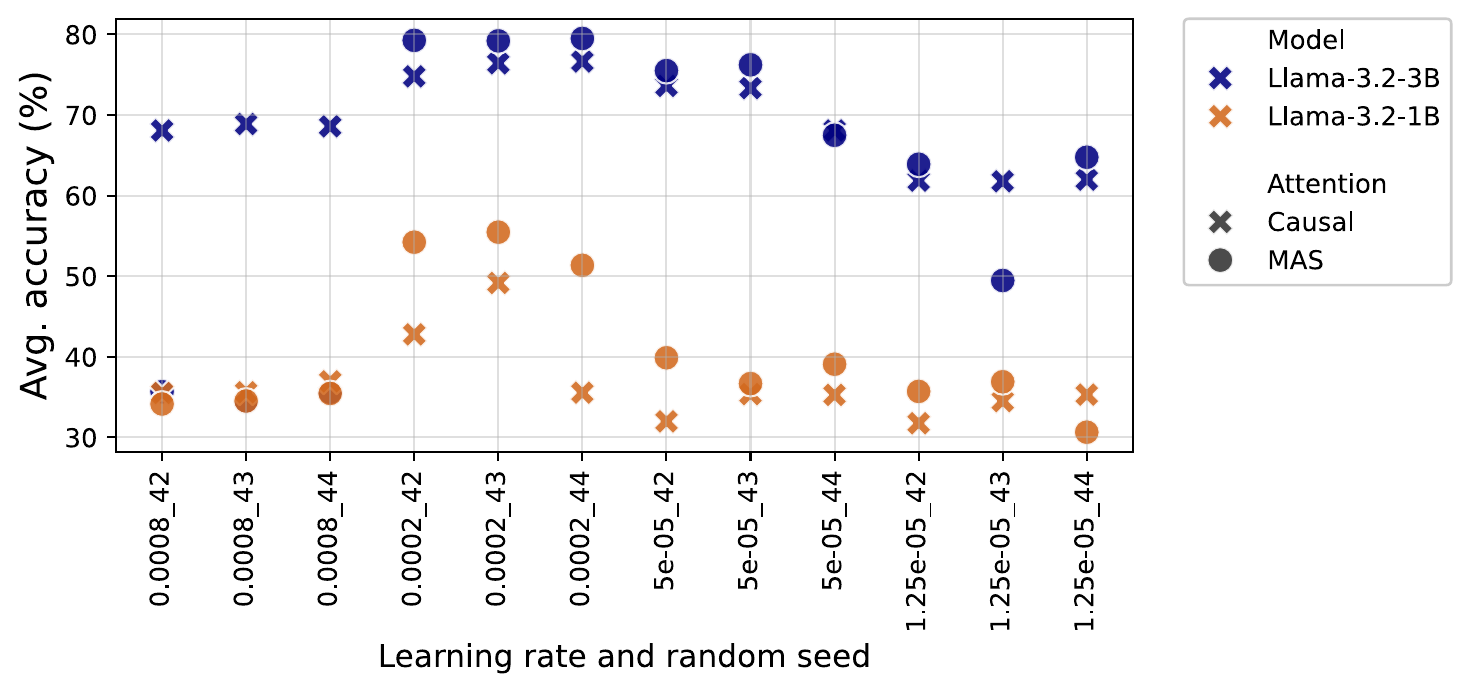}
  \caption{Average accuracy on 6 of the Commonsense Reasoning tasks, as a function of random seeds and learning rates.}
    \label{fig: lr X seed}
\end{figure*}

The results, shown in \autoref{fig: lr X seed}, confirm consistency across seeds, with slight variances for the lowest learning rates.
The optimal learning rate of approximately $1e^{-4}$ was consistent with both masking setups, indicating that MAS does not require a different learning rate range.
However, MAS demonstrated sensitivity to higher learning rates: beyond a certain threshold, performance degradation was more pronounced with MAS compared to causal masking.
This suggests that MAS requires more careful learning rate tuning to achieve optimal results.

\begin{figure*}[t]
\centering
\centerline{\includegraphics[width=1\columnwidth]{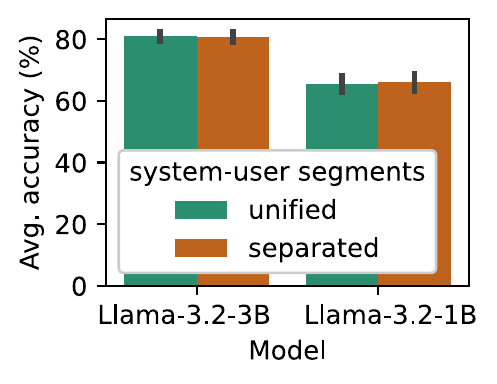}}
\caption{Average accuracy on the Commonsense Reasoning tasks for model with MAS, once where the system prompts and the user prompts are separated into two segments, and once when they are unified into one segment.}
\label{fig:system-user}
\end{figure*}

\subsection{System and Users Prompts Segments}
\label{System and Users Prompts Segments}
{In \autoref{Method}, we discussed our decision to separate the system prompt and the user prompt into two segments for computational efficiency. Here, we investigate whether this separation also affects the model's performance.}  

{In \autoref{fig:system-user}, we present the results of MAS's performance when the system and user prompts are combined into a single segment. 
The average accuracy across the Commonsense Reasoning benchmark show minimal differences compared to when the prompts are separated, suggesting that the separation does not significantly impact MAS's performance.}

{We hypothesize that the system prompt, being consistent across tasks, provides limited additional information for the model. Instead, the variability in the user prompt containing the actual question — is the primary driver of the model's performance.} 

Thus, separating the system and user prompts is justified not only by computational efficiency but also by empirical evidence, supporting its use as a practical and effective approach.

\section{MAS visualization}
\label{MAS visualization}  
\begin{figure*}[t]
\begin{center}
\centerline{\includegraphics[width=\linewidth]{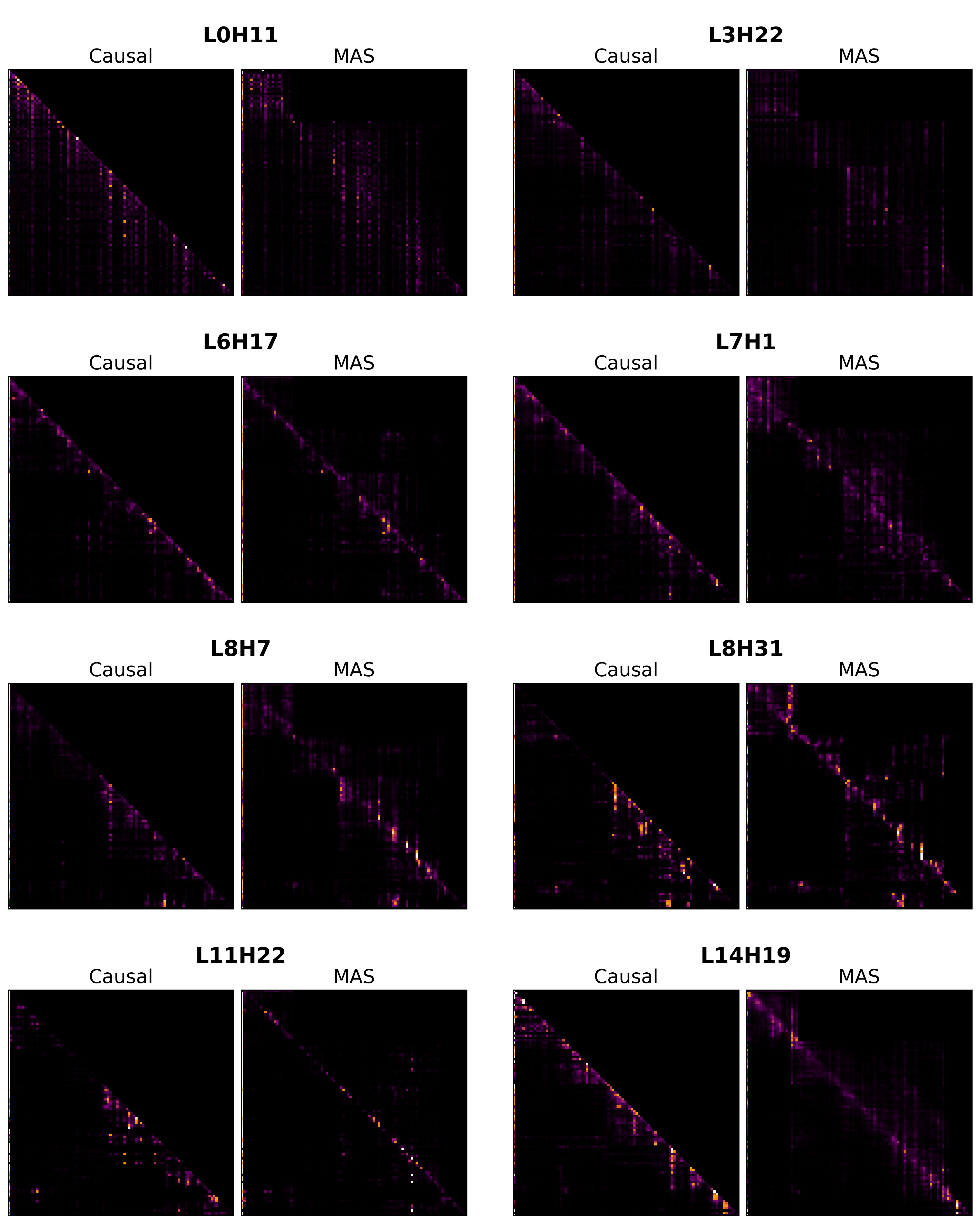}}
\caption{
The attention maps of two instances of fine-tuned Llama-3.2-1B, one with standard causal masking and the other with MAS
}
\label{fig: MAS viz appendix}
\end{center}
\end{figure*}

In \autoref{Interpretation}, we visualized attention maps for models fine-tuned with MAS.  
Here, \autoref{fig: MAS viz appendix} provides additional visualizations of two Llama-3.2-1B instances, one fine-tuned with standard causal masking and the other with MAS.  

Identifying the specific function performed by each attention head remains an active area of research \cite{zheng2024attention}, which lies outside the scope of this work.  
The visualizations are presented with two primary objectives:  
(i) To illustrate how MAS modifies attention patterns compared to standard causal masking.  
(ii) To provide a resource for future research aimed at gaining deeper insights into the behavior of models fine-tuned with MAS, as well as those trained using standard causal masking.

\end{document}